\documentclass[conference]{IEEEtran}
\IEEEoverridecommandlockouts
\usepackage{cite}
\usepackage{amsmath,amssymb,amsfonts}
\usepackage{algorithmic}
\usepackage{graphicx}
\usepackage{textcomp}
\usepackage{multirow}
\usepackage{xcolor}
\usepackage{booktabs}
\usepackage{authblk}
\usepackage{threeparttable}
\def\BibTeX{{\rm B\kern-.05em{\sc i\kern-.025em b}\kern-.08em
    T\kern-.1667em\lower.7ex\hbox{E}\kern-.125emX}}

\usepackage[hyperfootnotes=false, colorlinks=false,hidelinks]{hyperref} 
\usepackage[inline]{enumitem}
\usepackage[disable]{todonotes}

\DeclareMathAlphabet{\altmathcal}{OMS}{cmsy}{m}{n}
    
\begin{document}
\title{Scalable and Data-Driven Decision Support in the Maintenance, Repair, and Overhaul Process}
\author[1]{Houkun Zhu}
\author[1]{Helena Ebel}
\author[1]{Dominik Scheinert}
\author[1]{Florian Schmidt}
\author[2]{Jens Altenkirch}
\author[1]{Odej Kao}

\affil[1]{Technische Universit{\"a}t Berlin, Berlin, Germany}
\affil[2]{Siemens Energy Global GmbH \& Co. KG, Berlin, Germany}
\affil[ ]{Emails: \{h.zhu, firstname.lastname\}@tu-berlin.de, jens.altenkirch@siemens-energy.com}

% \author[1]{Anonymous}
% \author[2]{Anonymous}
% \affil[1]{Anonymous}
% \affil[2]{Anonymous}
% \affil[ ]{}
\maketitle

% conference
% https://www.ieem.org/public.asp?page=index.asp

% \section{Planned Content}
% \subsection{Background}
% Introduce MRO project; what it can offer to practitioners from engineers perspective and use cases.
% \subsection{Infrastructure}
% There are two options,
% 1. we describe the our own architecture only, which includes only Smart Expert architecture and ML pipeline.
% Pros: We have implemented most of them and fully under our control
% 2. We describe the archicture of the whole MRO project, which includes e.g., digital twin. As it is not in use, this might be hard.

% \subsection{evaluation}
% 1. pinpoint high explainability of our model
% 2. pressure test of our architecture
% 3. user study: compare the time consumption before Smart Expert and after;  Direct engineer feedback.

\begin{abstract}
Several businesses apply maintenance, repair, and overhaul (MRO) principles to the life-cycle of their existing products. 
In cases like casted gas turbine component Product Lifecycle Management (PLM), repairing components in frequent intervals can extend the lifetime expectation of the product, provide higher cost efficiency compared to newly produced components, and even improve the part design during the repair cycle. 
Another aspect of repair concerns sustainability, as products often contain rare materials. The emissions produced by the repair process are usually smaller than mining materials and casting new components. 

To optimize the repair process further, we propose the Smart Expert System (SES), which assists engineering experts with machine learning-based decision support throughout the repair process. 
We elaborate on its IT architecture and present machine learning models employed for representative MRO use cases.
The SES is evaluated using actual industry data from a leading gas turbine company and demonstrably fulfills formulated requirements concerning the suitability of the overall decision support and the stability of the enclosing IT architecture.
% In the evaluation, we conducted both qualitative results for the machine learning-based decision support and quantitative results showing strong performance and limitations of the given system architecture. Both evaluation scenarios were conducted with real industry data from a leading gas turbine company showing a significant acceptance of the SES.
%sustainability/repair instead of new build
\end{abstract}


\section{Introduction}
\label{sec:introduction}
% Some general MRO introduction
% - brief description of use-cases (what should the smart expert solve for the engineers?)

The fourth industrial revolution, also called Industry 4.0, is progressing exponentially, integrating horizontal and vertical manufacturing processes and linking products to improve industry performance and sustainability~\cite{dalenogare2018expected, ghobakhloo2020industry}. Many companies start with maintenance management in the Industry 4.0 practice because the transition from responsive and periodic maintenance to predictive and proactive maintenance is critical for companies to obtain technical and economical benefits\cite{prognosis2017mosyu}. Maintenance management is often referred to as Maintenance, Repair, and Overhaul (MRO) to describe the diverse cost-effective practice better. The MRO process is typically used in the context of gas turbine refurbishment to minimize life cycle costs and optimize the performance of the part.  

Availability and efficiency of gas turbine units are two main concerns for gas turbine operators regarding performance optimization~\cite{anton2007repair, pallos2001gas}. 
Many components have limited life expectations compared to the gas turbine unit's operational lifetime~\cite{pallos2001gas}. Consequently, maintenance and repair have always been in the interest of operators. The traditional procedure is usually executed by static repair process chains in which the same value-added steps are carried out continuously. 

However, the manual and static procedure is often inefficient. 
Typically, engineers refer to the Repair Specification Manual (RSM) to locate the corresponding repair actions or follow-up investigation process for an observed material defect. This procedure is monotonous, error-prone, and breaks the real-time capacity principle of documentation management in Industry 4.0\cite{GATTULLO2019276} due to manual search through the RSM documentation and infrequent RSM updates.
% Frequently arising unseen conditions are thus hindering effective use of RSMs as information is often outdated, while exceptional cases are not reflected in the RSMs manual.% case-by-case design and life evaluation investigations.

Additionally, component-specific knowledge---how to solve novel issues and execute unique or complex repair patterns---is built up continuously by individual engineering experts. In order to ensure a standardized and high-quality knowledge base, this knowledge needs to be documented and made available to all experts by the RSM.

To address the aforementioned challenges, a data-driven approach is effective because it can help automate the gas turbine repair process and boost efficiency in extracting and integrating new engineering expertise from unseen repairing operations data. Therefore, we introduce a scalable and data-driven assistant, Smart Expert System (SES), to support decision-making in the MRO process. Specifically, SES is applied in the following use cases we identified:
\begin{enumerate*}[series = tobecont, label=(\arabic*)]
    \item Automation of the repair decision process through digitalization and rule-based application of RSMs (Digital RSM),
    \item Inference of dispositions for defects not defined in RSMs through data-driven modeling of historical expert decisions (Quality Inspection),
    \item Estimation of the repair cost and overall feasibility of repair actions (Repair Scope).
\end{enumerate*} 

In this paper, we make the following contributions:
\begin{enumerate}
    \item We provide a machine learning-based decision support solution for the MRO process. 
    \item Our architecture can effectively handle varying demands through fine-granular system component scaling.
    \item Empirical results show that our solution can achieve high performance for repair operation predictions and effectively and reliably handle varying loads on the system via horizontal scaling. 
\end{enumerate}

The rest of this paper is organized as follows.~\autoref{sec:problem_statement} describes the problem statement and requirements for a digital assistant system.~\autoref{sec:analysis_pipelines} presents the general architecture design as well as solutions to each use case.~\autoref{sec:evaluation} evaluates the prediction accuracy of machine learning (ML) models in SES and scalability of SES.~\autoref{sec:related_work} introduces other intelligent assistant systems for Industry 4.0. 
In the end,~\autoref{sec:conclusion} concludes the paper and outlines future work. % 1 page together with abstract // Houkun

\section{Problem Statement}
\label{sec:problem_statement}

We first present the conventional MRO process and its associated manual decision-making steps. 
From this description, we then derive the concrete problems to be solved by SES and formulate further key requirements to be considered by the envisioned system.

%description of manual decision processes
\subsection{Conventional Engineering Process}
\label{subsec:conventional_engineering_process}
%TODO add image
\begin{figure}[h!]
    \centering
    \includegraphics{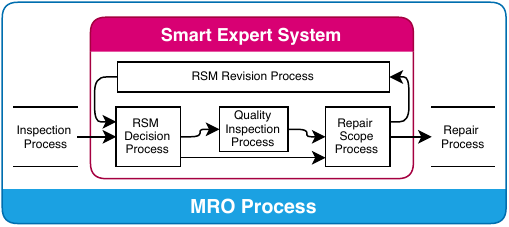}
    %\missingfigure{Workflow figure}
    \caption{MRO process and the corresponding use cases for the SES.}
    \label{fig:workflow}
\end{figure}
The conventional MRO process (see~\autoref{fig:workflow}) typically starts with the handover of a used machine component (e.g., a gas turbine blade) by a customer. The given component is inspected using different methods to detect all sorts of damages. Generally, the first step concentrates on an inspection phase, where all damages are documented to check the product's surface and material consistency by various tests, e.g., 3D scans. Based on the inspection results and decision rules defined in the RSM, an engineering expert decides on the severity of damage and possible repair options, including keeping or scraping the component.

As RSMs often focus on standardized damage patterns, rare defect types and complex defect patterns can be missed. Therefore, such defects require further inspection to determine a concrete repair process, i.e., a Quality Inspection (QI) process is started. 
Engineering experts manually inspect the component within the QI process and decide on a repair process based on their own experiences and expertise.

In order to learn and generalize from the cases handled via the QI process, the associated final repair decisions are documented within a Product Lifecycle Management (PLM) software.
%As last step, the machine has to be repaired and all decisions are documented throughout the process within PLM software.
Based on this knowledge, the engineering experts periodically adapt RSMs to keep the decision process documented. This standardization process usually occurs in longer intervals, commonly one or two years.

The outlined conventional engineering process bears potential for automation in various places, e.g., rule-based application of digital RSMs or inference of repair process decisions based on historical data. 
Hence, a digital assistant system is demanded. 

\subsection{Requirements for an Assistant System}
\label{sec:ses_requirements}
%Helena: Make requirements great again
In order to design and develop a valuable assistant system, we interviewed practitioners and collected user stories from the roles involved in the MRO Process. The gathered user stories indicate four core requirements for an assistant system.

(i) The system provides suggestions instead of automated decisions.
The main focus here is on using AI to assist the decision-making process. Engineers want to revise system results before passing on in the process. On the other hand, the system should support the incorporation of engineers' feedback on the prediction results to improve ML models.

(ii) The system has access to all decision-relevant data points and extensively incorporates them to assist in the MRO process.
% Completeness of all necessary information and expert knowledge for training ML models and automated prediction.
The system should be based on the MRO process's existing information and expert knowledge. For the training of ML models, it is necessary to process data, confirm intermediate results from experts, and create final results in a form that can be further applied in the MRO process and to link expert knowledge with the data sources.

(iii) ML models can adapt to new conditions.
Constantly emerging unseen conditions disqualify a static model. 
For example, it is conceivable that rare and previously unknown defect types arise or the data input changes due to technological advances, as a consequence exceeding defined limits. 
In order to continuously adapt to new conditions, the system should regularly retrain its models.

(iv) The reasoning behind system results is human-understandable.
To increase the acceptance of ML models, the system should provide engineers with sufficient explanations, e.g., regarding the decision-making process and the confidence of a particular model about generated predictions.
 % 1 page // Florian

\section{Analysis Pipelines}
\label{sec:analysis_pipelines}

This section describes the design and analysis pipeline of SES, which takes the problem statement described in~\autoref{sec:problem_statement} into account. 

%Dominik: Make images beautiful again

\subsection{Architecture Design}
\begin{figure}[h!]
    \centering
    \includegraphics[width=\columnwidth]{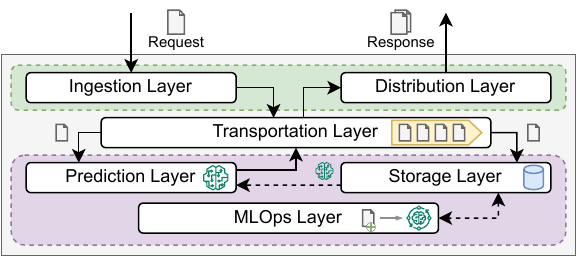}
    \caption{The architecture of SES. 
    The layers can be categorized with respect to 1) request handling and 2) modeling, with the transportation layer in between.}
    \label{fig:architecture}
\end{figure}  
The SES attempts to support engineers in various phases of the engineering process via intelligent and data-driven suggestions. 
Consequently, for its realization, we choose a distributed system, i.e., loosely coupled system components interacting with each other, which allows for scalability and maintenance of individual components. 
As shown in~\autoref{fig:architecture}, our architecture can be decomposed into six application layers:
\begin{enumerate}
    \item Ingestion Layer: The SES utilizes Restful APIs so that other systems can effortlessly integrate with the SES and provide data related to the engineering process.
    If required, the received data is further processed before being sent to the transportation layer.
        
    \item Transportation Layer: In order to facilitate communication among individual components and allow for asynchronous request handling and data processing, we employ a scalable message broker\footnote{\url{https://www.rabbitmq.com/}, Accessed: May 2022}. This distributed middleware component enables the publication and retrieval of data from message queues. To ensure a high level of performance, all data should be routed through this platform as it decouples components where slow sinks are regularly a limiting factor.
        
    \item Prediction Layer: The various prediction services are organized in this layer. 
    They consume data from message queues of interest, enrich it with respective predictions, and publish the transformed data again to suitable message queues.
    
    \item Persistence Layer: In order to retrain models when needed and keep track of provided user feedback, we persist requests, predictions, and potential correction of predictions reported by users.
    
    \item MLOps Layer: All ML operations are encapsulated by this layer. 
    This encompasses the training, updating, versioning, and serving of models.
    
    \item Distribution Layer: Once a user request has been processed and enriched by a model covering the respective use case, the transformed data is either 1) ready for retrieval by the initially requesting service, or 2) directly send there.
    
\end{enumerate}

The individual system components of this architecture are realized as containerized applications.
This allows for individual scaling of system components and thus efficiently adapting the SES to different demands, e.g. adapting to increased user requests or extending further ML models.
Using a container orchestration system such as Kubernetes\footnote{\url{https://kubernetes.io/}, Accessed: May 2022}, the deployment and management of our distributed system are further simplified.

\subsection{Use Cases}
To boost the conventional engineering process detailed in~\autoref{subsec:conventional_engineering_process}, we identify and provide solutions for the following use cases:

\subsubsection{Digital Repair Specification Manual}
The SES digitizes RSM to automate the defect evaluation and repair process. Specifically, the SES converts the existing RSM to a machine-readable form. Therefore, once the defect measurement data is available, engineers do not need to refer to the RSM, but acquire the appropriate repair operation recommendation directly.

\subsubsection{Quality Inspection}
In some cases, the defect situation is not defined in RSM. Engineering experts then need to investigate those defects and make repair decisions case by case. The expertise is generalized every one to two years and then integrated into the RSM. In order to boost the QI process, we employ ML to generalize previous experts' decisions and automate the QI. Specifically,  we choose decision tree models to learn the insights because they foster model explainability and thus enable trust in model predictions.

\subsubsection{Repair Scope}
Repair scope condenses information about all required repair operations assigned to a single part. It defines the severity level of defects and the complexity level of associated repair operations. It is used to calculate the repair cost and decide whether a component is worthwhile to repair. This is oftentimes rule-based but varies nonetheless, for instance across repair shop floors (technical equipment) or customers (financial budget and expectations). Therefore, our system digitizes the repair scope logic per shop floor and customer to automate this process.

\subsection{Fulfillment of Requirements}
Our proposed realization of the SES takes into consideration the requirements defined in~\autoref{sec:ses_requirements}.

With regards to requirement (i), the predictions produced by the SES can be visualized in a user interface and thus leave the final decision to the engineer in charge.
In case of unsuitable recommendations, an engineer can apply a correction, which in turn is sent to the SES using dedicated endpoints and used for fine-tuning the internally utilized machine learning models. 

The second requirement (ii) is demanding data completeness and data interpretability.
On the one hand, the SES is making use of existing documents such as the RSM for rule-based predictions, thereby making use of available expert knowledge.
On the other hand, the general problem modeling and close development of the SES together with engineers in the field assured that 1) the process is sufficiently incorporated and reflected, and 2) the utilized modeling techniques comply with the requirements of engineers regarding interpretability.%\todo{adapt to updated requirement (interpretability is not required in requirement 2}

Quickly adapting ML models are required following requirement (iii), and with our use of decision trees, the SES can adapt to new conditions through periodic retraining from scratch and fine-tuning of a warm-started model.

After the previous discussions, the last requirement (iv) regarding human-understandable results is considered fulfilled as well by the design of the ML model.
In this context, it is advisable to enrich an incoming request not only with the respective model prediction, but also with an indication of the certainty of the prediction produced. %\todo{adapt to updated requirement }

Besides the general fulfillment of requirements by design choices, we next look into the quantitative analysis of SES. %details about quantitative performance measurements.  % 1 page

\section{Evaluation}
\label{sec:evaluation}
In this section, we demonstrate experiment results of SES performance. Both the prediction performance and end-to-end latency are evaluated.

\subsection{Prediction Performance}

\begin{figure}[h!]
    \centering
    \includegraphics{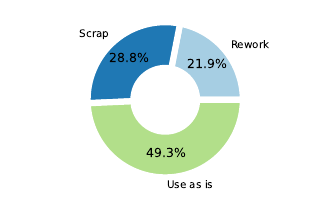}
    \caption{Imbalanced distribution in Quality Inspection training data.}
    \label{fig:qn_disposition_pie}
\end{figure}

To predict the disposition of a defect component in QI, we employ the decision tree model to provide a transparent and explainable model. There are three types of disposition decisions, \begin{enumerate*}[series = tobecont, label=(\arabic*)]
     \item Use as is: the component is good enough and demands no repair.
     \item Rework: the additional repair operations need to be conducted.
     \item Scrap: the component is beyond repair or worthless to repair.
 \end{enumerate*}
 As depicted in~\autoref{fig:qn_disposition_pie}, the training data is imbalanced, where half of the data are use as is. 
 
%  To cope with skewed class distribution, we use weighted metric to improve our 

It is critical to predict gas turbine component disposition conservatively in the industry because if the prediction is more optimistic than reality, it may bring potential risk to consumers' products, e.g., complete damage to a gas turbine. Therefore, it is more favorable to dispose components cautiously.

In order to handle skewed disposition distribution and conservative prediction, SES considers conservative preference ranks while making a decision, i.e., SES tends to make a more serious disposition prediction (overestimation) when the probability of that disposition is above a certain threshold. For example, if the threshold is 30\%, and the model outputs the probability of 40\% use as is, 22\% rework and 38\% scrap, conservative prediction will then select scrap as the prediction result because it ranks first and is greater than 30\%.

Proper evaluation metrics are essential to handle imbalanced data. SES uses F1 score to evaluate the prediction as it combines both recall and precision score. Furthermore, to quantify the cautiousness of our prediction, we define another metric, underestimation ratio, as,
\begin{equation*}
    \frac{\text{\# of underestimated prediction}}{\text{total predictions}}.
\end{equation*}
Underestimation is to predict use as is or rework for scrap, or to predict use as is for rework.

The evaluation data is based on real industry data from a leading gas turbine company, and the empirical results show a decrease of the underestimation ratio from 8.0\% to 5.0\% at the cost of 0.5\% drop in F1 score by utilizing the conservative prediction, i.e., SES can achieve 88.5\% F1 score with a 5.0\% underestimation ratio.

\subsection{End-to-End Latency}
We evaluate the end-to-end latency of SES in two settings. Firstly, all components in SES are running with two instances. Different pressure levels of requests are sent to simulate various loads. Secondly, we run 50 request threads and test the latency under a different number of scale-out, i.e., all components in our architecture are deployed with the same number of instances except the message broker. In both experiments, we measure the latency of requests in the first 60 seconds and run each setting 5 times. The experiments assess the processing speed and scalability of SES.

\subsubsection{Same Scale-out Under Different Loads}
\begin{figure}[h!]
    \centering
    \includegraphics[width=\linewidth]{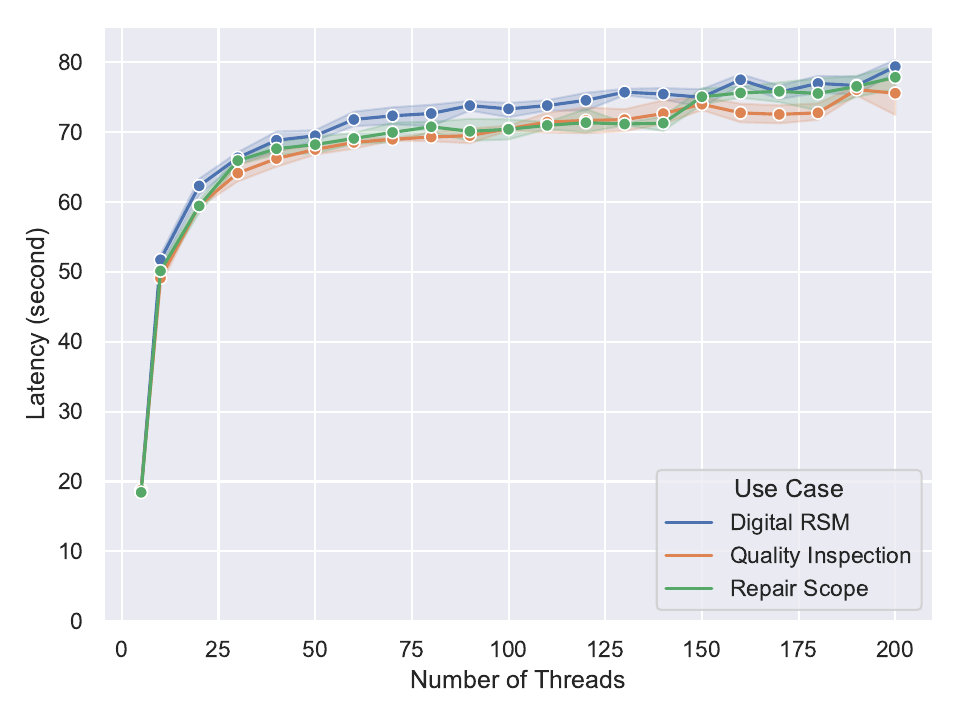}
    \caption{Latency under pressure test with the same scale-out (\# of instance is 2). The bigger the number of threads is, the greater load we generate.}
    \label{fig:pressure_test}
\end{figure}

In this experiment, we measure the latency of SES under different loads, i.e., we use a different number of request threads to simulate various loads. As presented in~\autoref{fig:pressure_test}, the latency rises significantly after increasing the pressure, i.e., latency surges from around 20 seconds to almost 80 seconds when we grow the number of request threads from 10 to 200.

\begin{figure}[h!]
    \centering
    \includegraphics[width=\linewidth]{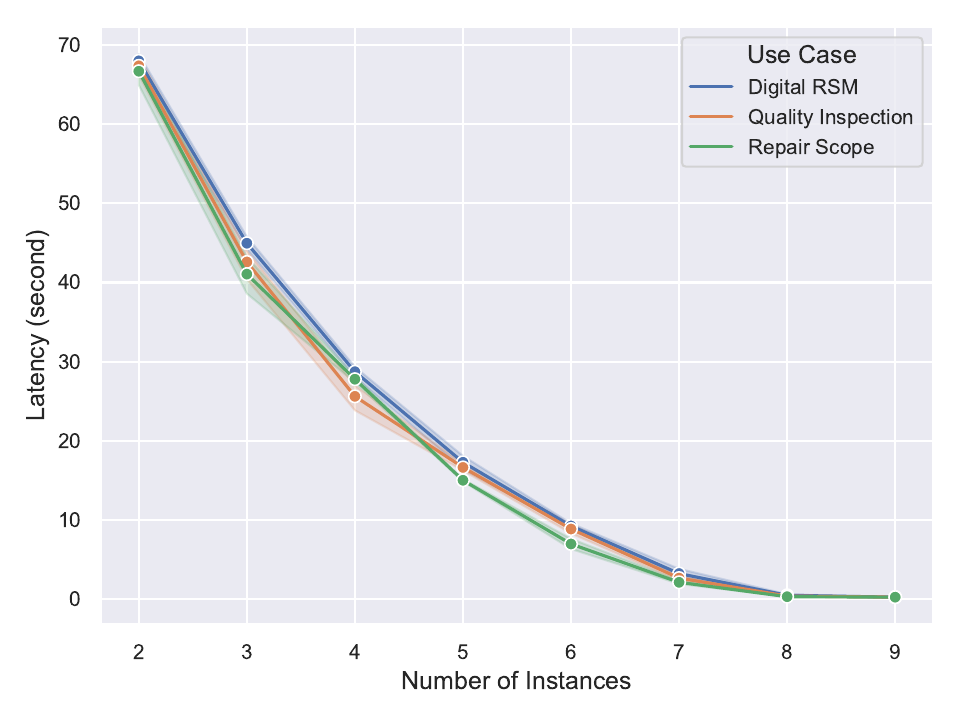}
    \caption{Scale out test under same load (\# of threads is 50). The higher number of instance we scale out, the lower latency we get.}
    \label{fig:scale_out_test}
\end{figure}

\subsubsection{Different Scale-out Under Same Loads}
In this experiment, we measure the latency of SES under a different number of scale-out, i.e., we use different numbers of SES instances. As illustrated in~\autoref{fig:scale_out_test}, scaling out of SES can noticeably decrease the latency.

Both experiments showed that the SES can effectively adapt to a surge in demand by scaling to ensure process efficiency.
 % 1 page

\section{Related Work}
\label{sec:related_work}

%Dominik: Make Related great again
This section presents related work from the general field of intelligent assistance solutions for Industry 4.0.

There exist related work for envisioned holistic solutions for intelligent, digital assistants.
%Other related work consider technologies other than AR, or envision a more holistic solution.
% (2019) Assistance systems for industry 4.0 environments (https://www.researchgate.net/profile/Bjoern-Barig/publication/331673690_Assistance_systems_for_Industry_40_environments/links/5c87b3c2a6fdcc88c39d47d8/Assistance-systems-for-Industry-40-environments.pdf)
% uses process-related “local and a global assistance system(s)”. The local assistants are associated with individual machine modules. They learn and analyse the production processes using Artificial Intelligence (AI) methods (neural networks etc.). The results from the local assistants are evaluated by the “global assistance system”. It has a central overview of the entire production chain. 
For instance, it is motivated in~\cite{barig2019assistance} that process-related assistance systems could be used both locally and globally, with the global assistance system evaluating individual results and thus having a central overview, whereas the local assistants are specialized on concrete machines or production processes.
% (2017) CASOA: An Architecture for Agent-Based Manufacturing System in the Context of Industry 4.0 (https://ieeexplore.ieee.org/stamp/stamp.jsp?arnumber=8053743)
% gives an overview of agent-based architectures for manufacturing systems. Besides, a cloud-assisted self-organized architecture is presented by comprising smart agents and cloud to communicate and negotiate through networks.
An agent-based manufacturing system is proposed in~\cite{TangLWD18}, with a cloud-assisted self-organized architecture designed for smart agents in the field, which makes it feasible for industry 4.0 and its typical edge environments.
% (2018) A Concept towards Automated Data-Driven Reconfiguration of Digital Assistance Systems (https://www.sciencedirect.com/science/article/pii/S2351978918304724?via%3Dihub)
% presents a concept for automated data-driven reconfiguration of digital assistance systems.
In order to increase the acceptance and lifespan of to-be-employed digital assistance, it is also important to follow a systematic approach to reconfiguration of such systems, which has been proposed in~\cite{REISINGER201899} and attempts to act as a concept for various common challenges.

% (2019) Survey: The evolution of the usage of augmented reality in industry 4.0 (https://iopscience.iop.org/article/10.1088/1757-899X/521/1/012017/meta)
% review / survey
Notably, a majority of works focuses on the usage of Augmented Reality (AR) to support various engineering tasks~\cite{Gallala2019}. 
% (2017) Smart operators in industry 4.0: A human-centered approach to enhance operators’ capabilities and competencies within the new smart factory context (https://www.mcmaster.ca/sept/learninghub/AMI4/pdfs/Smart-operators-industry4.pdf)
% "Sophos-MS": relies on Augmented Reality contents and on an intelligent personal digital assistant with vocal interaction capabilities
One example is Sophos-MS~\cite{LongoNP17}, a human-centered and modular solution which relies on AR contents and on a personal digital assistant with vocal interaction support.
% (2021) An Augmented Reality inspection tool to support workers in Industry 4.0 environments (https://www.sciencedirect.com/science/article/abs/pii/S0166361521000191)
% AR inspection tool to support workers at the workplace in Industry 4.0 contexts, users can easily detect design discrepancies and add 3D notes to 3D models
Another example is presented in~\cite{MarinoBCFB21}, an AR inspection tool that allows detection of discrepancies on 3D models and subsequent annotation of those, effectively improving problem identification and digital documentation.
Similarly, a methodology is proposed in~\cite{GATTULLO2019276} for the conversion of analog documentation manuals and for the authoring of new AR-based manuals in accordance with Industry 4.0 principles.
% (2018) Augmented Reality for Industry 4.0: Architecture and User Experience (http://ceur-ws.org/Vol-2299/paper5.pdf)
% review + two uses cases + three main requirements + recommendations
On a meta-level, the authors of~\cite{JaklSHW18} generally review the state of the art for AR in industry, and work out architecture considerations and user experience recommendations based on use cases provided by cooperating Industry 4.0 companies.

More recently, researchers started to review existing approaches in this field to derive general recommendations and reveal current limitations. 
% (2021) Worker assistance systems in manufacturing: A review of the state of the art and future directions (https://www.sciencedirect.com/science/article/abs/pii/S0278612521000534)
% Systematic literature review of worker assistance systems in production. This information is then used to derive strategic measures addressed to stakeholders for either exploiting the benefits of such systems, for conducting further basic and applied research or for improving the applicability and potential for industry.
For instance, a systematic literature review of worker assistance systems in production has been conducted by the authors in~\cite{MARK2021228}, aiming at classification of existing systems, discussion of the associated user groups and their individual requirements, as well as motivation of future research directions.
% (2021) Anatomy of a Digital Assistant (https://link.springer.com/chapter/10.1007/978-3-030-85910-7_34)
% presents an overview of a preliminary catalog of benefits in manufacturing. It covers central access, customization, delegation and guidance, eyes-free and hands-free interactions, mobile assistance, the support of multiple interface types, permanent accessibility, and speed.
Similarly, the general benefits of using digital assistance systems in manufacturing are discussed in~\cite{WellsandtHT21}, thereby among other things touching upon considerations of accessibility, speed, customization, and interaction capabilities.

Evidently, the aforementioned works mainly attempt to reinvent the interaction process of various manufacturing tasks, consequently demanding for a holistic approach.
With the SES solution and architecture proposed in this paper, we focus more on supporting engineers in the decision-making process in a data-driven manner through interactive suggestion of options.

 % 2/3 page together with conclusion // Dominik

\section{Conclusion}
\label{sec:conclusion}

In this paper, we presented SES, a scalable and data-driven system to assist gas turbine repairing process. It automates the conventional engineering process in MRO. Furthermore, it assists engineering experts with ML-based decision support in defects undefined in the RSM. Lastly, SES can effectively adapt to different demand and meet latency requirements due to its scalable architecture.
In our experiments, we evaluated the performance of SES. Empirical results showed that SES can achieve 88.5\% F1 score and 5.0\% under-estimation ratio. Additionally, the SES is able to mitigate and even prevent latencies from increasing by scaling out.

In the future, we want to evaluate and strengthen the transferability of our model on different repair objects. % 1/6 - 1/4 page

\section*{Acknowledgment}
This work has been supported through grants by the German Federal Ministry of Education and Research (BMBF) as MRO 2.0 which is co-financed by the European Regional Development Fund (ERDF). 

\bibliographystyle{IEEEtran}
\bibliography{bib}

% \section*{Glossary}
% \todo[inline]{delete this section before submission.}

% According to Jens, we can freely use all terms as Siemens' counterparts use them as well. The only thing we need to avoid is to use absolute value in measurement since it may leak design information. Otherwise it is not an issue for Siemens. 
% The below table is used to keep consistency of terms used in this paper. It will be removed before paper submission.
% \begin{table}[h!]
% \centering
% \begin{tabular}{ll}
% \toprule
% Term    & Translated Term \\
%  \midrule
% TLV & repair rule catalog\\
% Digital TLV &  digital repair rule catalog                \\
% FPI         & a procedure to find visual defects                 \\
% DDC & all measurement  \\
% QN          &  quality inspection               \\
% Smart Expert & Smart Expert System (SES)  \\ \bottomrule
% \end{tabular}
% \end{table}

\end{document}